\documentclass[sn-basic]{sn-jnl}% Basic Springer Nature Reference Style/Chemistry Reference Style

\usepackage{graphicx}%
\usepackage{multirow}%
\usepackage{amsmath,amssymb,amsfonts}%
\usepackage{amsthm}%
\usepackage{mathrsfs}%
\usepackage[title]{appendix}%
\usepackage{xcolor}%
\usepackage{textcomp}%
\usepackage{manyfoot}%
\usepackage{booktabs}%
\usepackage{algorithm}%
\usepackage{algorithmicx}%
\usepackage{algpseudocode}%
\usepackage{listings}%
\usepackage{hyperref}
\theoremstyle{thmstyleone}%
\theoremstyle{thmstyletwo}%

\theoremstyle{thmstylethree}%
\newcommand{\colorize}[2]{\colorbox{red!#1!white}{\strut #2}}

\begin{document}

\title[Article Title]{Enhancing and Accelerating Large Language Models via Instruction-Aware Contextual Compression}


\author*[1]{\fnm{Haowen} \sur{Hou}}\email{houhaowen@gml.ac.cn}
\equalcont{These authors contributed equally to this work.}

\author[1]{\fnm{Fei} \sur{Ma}}
\equalcont{These authors contributed equally to this work.}

\author[1]{\fnm{Binwen} \sur{Bai}}

\author[1]{\fnm{Xinxin} \sur{Zhu}}

\author[1]{\fnm{Fei} \sur{Yu}}

\affil*[1]{\orgname{Guangdong Laboratory of Artificial Intelligence and Digital Economy (SZ)}, \orgaddress{\street{Yutang}, \city{Shenzhen}, \postcode{518000}, \state{Guangdong}, \country{China}}}


\abstract{Large Language Models (LLMs) have garnered widespread attention due to their remarkable performance across various tasks. However, to mitigate the issue of hallucinations, LLMs often incorporate retrieval-augmented pipeline to provide them with rich external knowledge and context. Nevertheless, challenges stem from inaccurate and coarse-grained context retrieved from the retriever. Supplying irrelevant context to the LLMs can result in poorer responses, increased inference latency, and higher costs. This paper introduces a method called Instruction-Aware Contextual Compression, which filters out less informative content, thereby accelerating and enhancing the use of LLMs. The experimental results demonstrate that Instruction-Aware Contextual Compression notably reduces memory consumption and minimizes generation latency while maintaining performance levels comparable to those achieved with the use of the full context. Specifically, we achieved a 50\% reduction in context-related costs, resulting in a 5\% reduction in inference memory usage and a 2.2-fold increase in inference speed, with only a minor drop of 0.047 in Rouge-1. These findings suggest that our method strikes an effective balance between efficiency and performance.}




\maketitle

\begin{figure}
  \includegraphics[width=\textwidth]{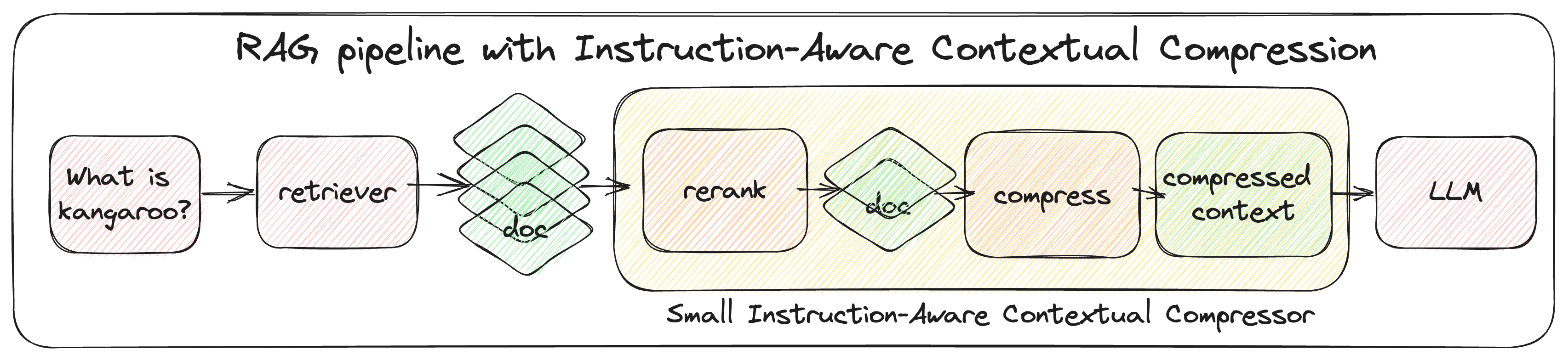}
  \caption{Retrieval Augmented Generation(RAG) pipeline with Instruction-Aware Contextual Compression.}
  \label{fig:teaser}
\end{figure}

\section{Introduction}\label{sec1}

Large language models (LLMs) have exhibited impressive capabilities in terms of both their robust performance and generalization across a diverse spectrum of natural language processing tasks, as well as practical real-world applications~\citep{brown2020language, touvron2023llama, touvron2023llama2}. 
To address certain issues with Large Language Models (LLMs), such as long-context~\citep{Xu2023RetrievalML} or hallucination~\citep{Ji2022SurveyOH, Shuster2021RetrievalAR} problems, retrieval-augmented generation (RAG)~\citep{lewis2020retrieval} has emerged. RAG has become an important approach to enhance  Large Language Models.

However, when using RAG, there can still be problems with irrelevant information. On one hand, inaccurate recall may lead to the retrieval of irrelevant documents. On the other hand, even within relevant documents, there might be irrelevant content that could distract the Large Language Model (LLM) from the relevant information. Passing the full document to the LLMs can lead to poor responses, large inference latency, and high costs.

Contextual compression aims to address this issue. The concept is straightforward: rather than directly presenting retrieved documents in their original form, they can be compressed, ensuring that only relevant information is conveyed.
Some research efforts~\citep{Li2023CompressingCT} have been committed to effectively compressing the context or prompt for large language models, with the aim of utilizing the most concise input while simultaneously preserving the robust performance of these models. 

In this paper, we introduce \textit{Instruction-Aware Contextual Compressor} (IACC), a novel approach that harnesses both ranking and generation information to eliminate extraneous context, thereby mitigating the computational overhead associated with the given context. Consequently, this leads to a reduction in both inference memory usage and inference time. The Instruction-Aware Contextual Compressor is adept at preserving finely detailed content directly related to instructions while compactly representing context, resulting in an efficient and streamlined input for Large Language Models (LLMs) without compromising their performance.

The main contributions of our paper are as follows: 
\begin{enumerate}
    \item We introduce Instruction-Aware Contextual Compressor, an innovative model aimed at enhancing the context efficiency of LLMs, which is able to reduce memory usage and inference latency without sacrificing LLMs performance. 
    \item We found that Instruction-aware contextual compression by generation is more effective than Instruction-aware contextual compression by ranking, even though the former utilizes training data that is only one-tenth of the latter.
    \item We developed the WikiQA-LongForm Dataset, a long-form open-domain question answering dataset based on Wikipedia entries, which can be used for training and evaluating models' context compression capabilities. This dataset is now publicly available at \href{https://huggingface.co/datasets/howard-hou/WikiQA-LongForm}{WikiQA-LongForm} for use in other research projects.
\end{enumerate}

The remaining sections of the paper are structured as follows: In Section 2, we delve into the related work. Section 3 outlines the method and model architecture. Section 4 describes experimental setup. Results are detailed in Section 5, and we provide conclusion in Section 6.

\section{Related Work}\label{sec2}

In this section, we review existing approaches~\citep{liu2022makes,lu2022fantastically,honovich2022instruction,wei2022chain} aimed at addressing the limitations imposed by context length in Large Language Models (LLMs). These limitations have motivated the development of various techniques to extend the context window of LLMs and enhance their performance.

\subsection{Retrieval-Augmented Generation}
Retrieval has been integrated into language models for years to enhance various aspects such as perplexity~\citep{borgeaud2022improving, wang2023shall}, factual accuracy~\citep{nakano2021webgpt}, downstream task accuracy~\citep{guu2020retrieval, izacard2021leveraging, izacard2022few, lewis2020retrieval}, and in-context learning capability~\citep{huang2023raven}. Combined with a standalone retriever~\citep{karpukhin2020dense, wang2022text, lin2023train}, retrieval-augmented LLM is a well-established approach for addressing question answering with long documents in an open-domain context.
In previous studies, language models have been augmented with retrieval during inference~\citep{khandelwal2019generalization, yogatama2021adaptive}, fine-tuning~\citep{izacard2022few, lewis2020retrieval, guu2020retrieval}, and pretraining~\citep{borgeaud2022improving, izacard2022few, wang2023shall}.
There are also some methods that aim to integrate LLM and retriever into a single model, creating an end-to-end solution \citep{jiang2022retrieval, shi2023replug}. 

\subsection{Long Context Large Language Models}
Many approaches have sought to improve the handling of longer contexts in Large Language Models (LLMs) through modifications to their underlying architectures. Notably, the Longformer~\citep{beltagy2020longformer} employs a linear attention mechanism that scales with sequence length, allowing it to accommodate longer contexts effectively. CoLT5~\citep{ainslie2023colt5} introduces conditional computation techniques that enable the model to focus more on crucial tokens in both feedforward and attention layers. However, it's worth noting that many existing works have not yet adopted such architectural modifications, mainly due to the high computational cost associated with training LLMs.
Another category of approaches addresses the context length limitation by employing context chunking strategies. The Parallel Context Windows~\citep{ratner2023parallel}  proposes a parallel context window method, which calculates attention within individual chunks and incorporates positional embeddings between these chunks.

\subsection{Prompt Engineering}
% 模型架构这个切入点不太好，咱们做的事情和模型创新关系不大，主要还是一种prompt engineering。建议从prompt engineering入手，写一写最新提出的prompt engineering的方法有哪些，和咱们这个做context crompression的联系与区别
Prompt engineering is a relatively emerging discipline focused on crafting and refining prompts to harness the power of language models (LMs) for diverse applications and research endeavors. Proficiency in prompt engineering aids in gaining a deeper insight into the capabilities and constraints of large language models (LLMs).
Researchers employ prompt engineering to enhance the performance of LLMs across an array of common and intricate tasks, including question answering and arithmetic reasoning. Developers leverage prompt engineering to devise resilient and efficient prompting strategies that interact seamlessly with LLMs and other associated tools.
Prompt engineering encompasses two key directions: text-to-text and text-to-image~\citep{wang2023review,oppenlaender2022prompt} interactions. This area of research has witnessed significant manual efforts, exemplified by A Prompt Pattern Catalog~\citep{white2023prompt}, where a comprehensive collection of handcrafted prompt techniques is meticulously documented. On the other hand, there are automated approaches to prompt generation, such as the work on "Automatic Prompt Engineer" by Zhou and "LM-BFF" by Gao~\citep{zhou2022large,gao2021making}.

\subsection{Context Compression}
Context compression can be considered a form of prompt engineering, although their emphases are slightly different. Several techniques aim to compress prompts effectively while maintaining context relevance. The Selective Context~\citep{Li2023CompressingCT} approach leverages concepts from information theory, specifically self-information, to compress the context. Another approach, Learning to Compress Prompts with Gist Tokens~\citep{mu2023learning}, trains Gist models to compress prompt words into "gist" tokens before inputting them into the LLMs. Additionally, LeanContext~\citep{arefeen2023leancontext} extracts a dynamic number $k$ of key sentences from prompts and uses a reinforcement learning mechanism to determine the optimal value of $k$ for compression. 

\section{Method}
\begin{figure*}[t]
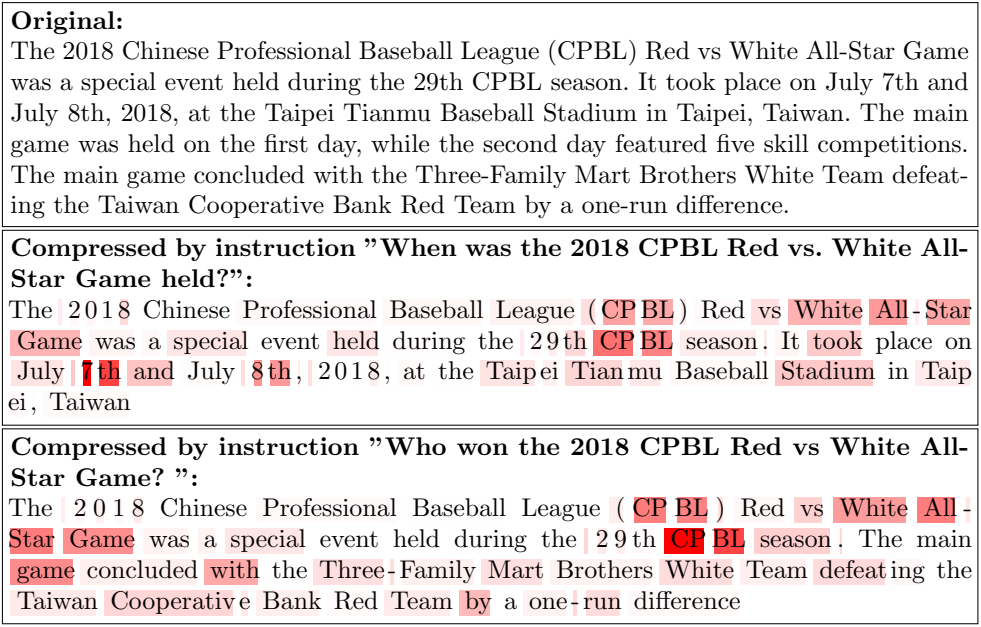

    \centering
\resizebox{\textwidth}{!}{
    \fbox{\parbox[c]{\textwidth}{ 
\textbf{Original: }{\setlength{\fboxsep}{-1pt}
 \input{doc}
}
   \par }}
}
\resizebox{\textwidth}{!}{
\fbox{\parbox[c]{\textwidth}{
\textbf{Compressed by instruction "When was the 2018 CPBL Red vs. White All-Star Game held?": } {\setlength{\fboxsep}{-1pt}
 \input{case1}
}
    }}
}
\resizebox{\textwidth}{!}{
\fbox{\parbox[c]{\textwidth}{
\textbf{Compressed by instruction "Who won the 2018 CPBL Red vs White All-Star Game? ":} {\setlength{\fboxsep}{-1pt}
 \input{case2}
}
    }}
}
    \caption{A visualisation of Instruction-Aware Contextual Compression. Deeper color indicates a stronger relevance to the instruction.}
    
    \label{fig:color_text}
    
\end{figure*}

We introduce Instruction-Aware Contextual Compression~\citep{Li2023CompressingCT}, an innovative approach for context compression that leverages both ranking and generation information. 
In contrast to the instruction-agnostic context compression methods used previously, Instruction-Aware Contextual Compression is a method that relies on instructions to perform context compression. Depending on the specific instruction provided, the model produces different compression outcomes, as shown in Figure \ref{fig:color_text}, removing irrelevant portions of the context, ultimately achieving improved context compression results.

To achieve Instruction-Aware Contextual Compression, we propose a two-stage pre-training methodology comprising the following stages:
(1) Ranking-Based Learning Stage.
(2) Generative Learning Stage. This section commences with an exposition of the model architecture employed in Instruction-Aware Contextual Compressor, and then introduces how we trained it in two different stages.

\subsection{Model Architecture}

\begin{figure}[h]
  \centering
  \includegraphics[width=\linewidth]{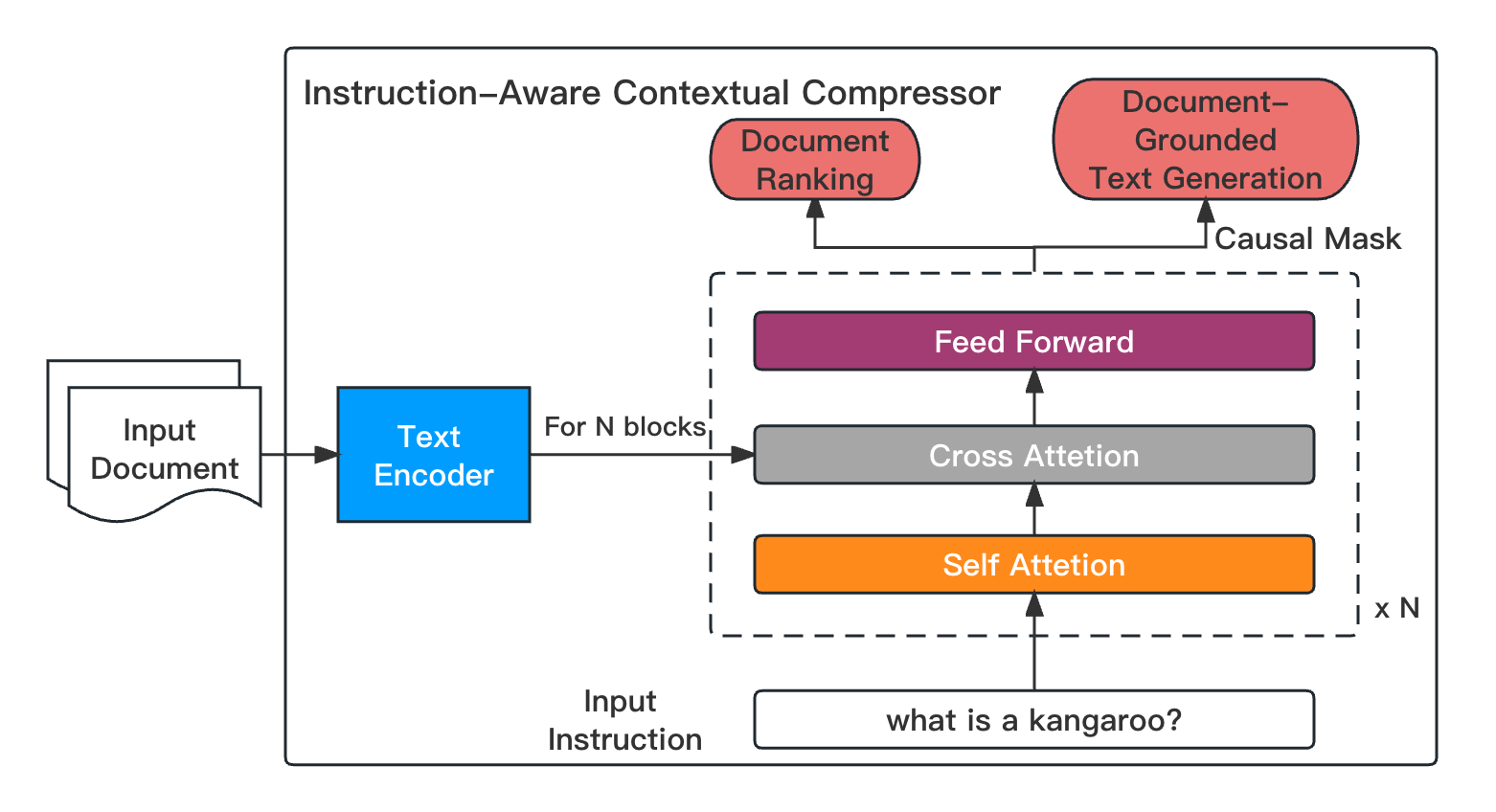}
  \caption{Model architecture of Instruction-Aware Contextual Compressor. We jointly optimize two objectives which enforce the model to extract contextual representation most relevant to the instruction.}
  \label{fig:arch}
\end{figure}

We introduce Instruction-Aware Contextual Compressor as a trainable module to implement the Instruction-Aware Contextual Compression method. Instruction-Aware Contextual Compressor adopts an encoder-decoder architecture, consisting of both an encoder and a decoder, as illustrated in Figure~\ref{fig:arch}.

The document encoder is a standard multi-layer transformer encoder and utilized to extract features from the input documents.

The decoder is a standard multi-layer transformer decoder, which equipped with two distinct functionalities, which can be toggled by modifying the masking:

\begin{enumerate}
    \item The ranking decoder performs re-ranking based on the output features obtained from the document encoder. In this mode, interaction occurs between instruction features and document features within cross-attention layers. The decoder employs bidirectional self-attention layers without any masking.
    \item The generation decoder, on the other hand, replaces the bidirectional self-attention layers in the decoder with causal self-attention layers. It uses a [BOS] token to denote the start of a sequence and an end-of-sequence token to signify its conclusion.
\end{enumerate}

The model consists of 8 encoder and 8 decoder layers, which affect its depth and ability to capture complex patterns. The primary model dimension is 512 and it uses a feed-forward dimension of 1024 for its inner layers. The model has 6 attention heads, allowing it to focus on different aspects of the input data. 
The model has a total of 0.18 billion trainable parameters, which is significantly smaller than the current mainstream large language models like Llama ~\citep{touvron2023llama, touvron2023llama2}, which have 7 billion, 13 billion, and 70 billion parameters, respectively. 
We initialize Instruction-Aware Contextual Compressor with the pre-trained weights of umT5\cite{Chung2023UniMaxFA}, which has been pre-trained on a multilingual corpus, enabling it to handle multilingual tasks effectively. 

The model's architecture, in conjunction with its training objectives, empowers it to capture the intricate interplay between instructions and documents, facilitating the extraction of the most pertinent information from the documents.

\subsection{Training Objectives}

We jointly optimize two objectives during training,
with one ranking-based objective and one generation based objective. Each instruction-document pair only requires one forward pass through the document encoder, and two forward passes through the decoder, where different functionalities are activated to compute the two losses as delineated below

\textbf{Ranking Loss} activates the ranking decoder. It aims to learn instruction-document representation that captures the fine-grained alignment between instruction and document. In ranking task, the model uses a ranking head (a linear layer) to predict a instruction-document matching score given their instruction feature and document feature. All positive samples are placed in the first position, and an additional 19 hard negative samples are retrieved by the retriever. Then, the model is trained as a 20-class classification task. This approach effectively boosts the scores of positive samples while suppressing the scores of negative samples. The specific formula for calculating the loss is as follows:

\begin{equation}
L_{ranking} = -\sum_{i}^{c} t_i log(\frac{e^{s_i}}{\sum_{j}^{c}{e^{s_j}}})
\end{equation}

where $s_i$ represents the ranking score of $i$-th sample. $t_i$ is the $i$-th target. $c$ represents the number of class.

\textbf{The Language Modeling Loss} activates the generation decoder, with the goal of generating useful response based on the provided context and instruction. It is optimized using a cross-entropy loss that trains the model to maximize the likelihood of the text in an autoregressive manner. We incorporate a label smoothing factor of 0.1 when calculating the loss. In comparison to the ranking loss, the Language Modeling loss equips the model with the ability to generalize for  following instructions. This empowers the model to gain a deeper understanding of the potential correct context location and to effectively model fine-grained correlations. the language modeling loss can be written as:

\begin{equation}
L_{lm} = -\sum_{t=1}^{N} \log{p(y_t | y_{<t})}
\end{equation}
where $p(y_t | y_{<t})$ denotes the output probability for the correct token $y_t$ given the previous context $y_{<t}$. $N$ denotes the sequence length. 

\subsection{Instruction-Aware Contextual Compression by Ranking}

Instruction-Aware contextual compression by ranking is a fairly straightforward process. First, an appropriate text-splitting strategy is employed, which can involve splitting based on specific character or by length, effectively converting the document into multiple chunks. Next, the model's ranking capability is applied to score chunks with instruction, followed by sorting them. Chunks are then retained based on a specified percentage. Throughout the compression process, care is taken to maintain the order of the chunks.

\subsection{Instruction-Aware Contextual Compression by Generation}

Compressing context by generation leverages the ability of Grad-CAM~\citep{Selvaraju2016GradCAMVE} to capture fine-grained relevance. 
The calculation process of Grad-CAM can be summarized as follows: Firstly, perform a forward pass to obtain the final classification probabilities. Calculate the gradients of the token with respect to the target class. Average the gradients for each token and extract an attention map within a specific cross-attention layer. Grad-CAM scores can be obtained by multiplying the attention map and the gradient weight vector. These Grad-CAM scores can be considered as the contribution of each token to the classification result. 

There's no need to pre-split the context; instead, the entire context is input into the model. Following that, k-step responses are generated based on the context and instruction. In the Instruction-Aware Contextual Compressor, attention maps and gradients are recorded for specific cross-attention layers, which are used to compute token-level Grad-CAM scores. These token-level Grad-CAM scores are then averaged to obtain chunk/sentence-level Grad-CAM scores. The chunks are subsequently sorted based on their Grad-CAM scores, and a specified percentage of the highest-scoring chunks is retained. Similarly, throughout the compression process, care is taken to maintain the order of the chunks.

\subsection{Ensemble the two methods}

Effectively ensemble these two methods of Instruction-Aware contextual compression can yield better compression results. The magnitude difference between the ranking score and Grad-CAM score makes it challenging to determine a suitable weighting parameter for fusion. Therefore, in the end, we opted for a non-parametric approach. Specifically, we individually rank the two types of information and then use the average of the two rankings to compress the context.

\section{Experiments}

\subsection{Datasets}
% 数据集分两部分，一部分是用于排序的预训练数据；一部分是用于问答训练的数据。
In this study, we utilized two types of datasets. The first are ranking datasets, designed to empower the model with robust re-ranking capabilities. The second are generation datasets, intended to equip the model with generative abilities. The ranking datasets comprise 15 million samples, while the generation datasets consist of 1.63 million samples. 

\subsubsection{Ranking datasets}

\textbf{$\mathbf{T^2}$Ranking~\citep{Xie2023T2RankingAL}} is a large-scale Chinese passage ranking dataset published in April 2023, which comprises 307K queries and 2.3M unique passages from real-world search engines. To constructing more accurate ranking algorithms, each query-passage pair has 4-level fine-grained annotations. For the retrieval task, it classifies Level-2 and Level-3 passages as relevant passages, while categorizing all remaining passages as irrelevant.

\textbf{M3E Dataset}~\citep{m3e} comprises a total of 22M sentence pair samples from a diverse range of topics, including Chinese encyclopedia, finance, healthcare, law, news, academia, etc.
This dataset mainly consists of datasets used for other tasks, among which over 3M of data is instruction fine-tuning data, while some datasets comes from tasks such as Q\&A, parallel semantics, machine reading comprehension, corpus, NL2SQL, text classification, text summarization, natural language processing, etc.

\subsubsection{Generation Datasets}

\textbf{Dureader Dataset~\citep{He2017DuReaderAC}}  is a extensive open-domain Chinese machine reading comprehension dataset, encompassing 200,000 questions, 420,000 answers, and 1 million documents. The questions and documents are sourced from Baidu Search and Baidu Zhidao, while the answers are manually crafted. In this study, we only use its robust subset, which contains 14,500 samples for training and 1.42k samples for validation.

\textbf{WikiQA-LongForm Dataset} is a long-form open-domain question answering dataset based on Wikipedia entries. We employed a heuristic approach and our proprietary NLU model to filter out lower-quality and sensitive or controversial entries, retaining 254,547 high-quality entries. These entries were transformed into multi-turn dialogue data using ChatGPT, resulting in high-quality Long Form QA data after further heuristic filtering. The WikiQA-LongForm Dataset is a contribution of this study and is publicly available at \href{https://huggingface.co/datasets/howard-hou/WikiQA-LongForm}{WikiQA-LongForm} for use in other research projects.

\subsection{Large Language Models}

During our experimentation, we conducted tests using Instruction-Aware Contextual Compressor on ChatGPT, which is based on the GPT-3.5-turbo-0613 architecture. ChatGPT represents an Instruct-tuned language model that has undergone further enhancement through Reinforcement Learning from Human Feedback (RLHF) and boasts an impressive 175 billion parameters. The foundational language model of ChatGPT appears to be code-davinci-0022, and previously, davinci, as outlined in \cite{brown2020language}. Our objective was to compare the performance of ChatGPT with and without the application of Instruction-Aware Contextual Compressor to gain insights into its impact on the model's efficiency and accuracy. 
The settings of ChatGPT are all set to their default values, except for the $top\_p$ parameter, which is set to 0.1. This adjustment is made to reduce the impact of randomness on the evaluation results.

\subsection{Experimental Settings}

We conduct a comparative analysis to assess the effectiveness of Instruction-Aware Contextual Compressor and analyze the associated trade-offs.

\textbf{RAG Baseline}: The table \ref{tab:rag_baseline} displays several RAG baselines, including scenarios where RAG is not used, direct feeding of the correct context to the large language model, the scenario where only a retriever is used in pipeline, and the scenario where a retriever is used in combination with Instruction-Aware Contextual Compressor for re-ranking.

\begin{table}[htbp]
    \centering
    \caption{The LLM performance in different scenarios}
    \label{tab:rag_baseline}
    \begin{tabular}{lcccc}
        \toprule
        & Rouge-1 & Rouge-2 & Rouge-L & Recall@1 \\
        \midrule
        Ground Truth & 0.683 & 0.539 & 0.631 & 1.0 \\
        Recalled Top1 & 0.656 & 0.507 & 0.605 & 0.86 \\
        Recall + Rerank Top1 & 0.675 & 0.529 & 0.623 & 0.962 \\
        \bottomrule
    \end{tabular}
\end{table}

\textbf{Compression Baseline}: Our evaluation involves a comparison between Instruction-Aware Contextual Compressor and Selective Context~\citep{Li2023CompressingCT}, which employs a basic approach to filter out an equivalent amount of data based on self-information. Selective Context utilizes the GPT-2 model with 124 million parameters.
Some readers may doubt whether the GPT-2 124M model is too small to be considered a sufficiently robust baseline for comparison.
To this end, we used Baichuan-7B with 7 billion parameters to run the Selective Context, which has parameters 56 times larger than GPT-2.
The results at a retention ratio of 0.5 are presented in the table below:

\begin{table}[h]
\centering
\caption{Comparison of ROUGE-L scores for the Selective Context method using Baichuan-7B and GPT-2 models.}
\begin{tabular}{lcccc}
\hline
Model & Parameters & R (rouge-l) & P (rouge-l) & F (rouge-l) \\
\hline
Baichuan-7B & 7B & 0.5448 & 0.5419 & 0.5024 \\
GPT-2 & 124M & 0.5758 & 0.5506 & 0.5205 \\
\hline
\end{tabular}
\label{tab:rouge-comparison}
\end{table}

We found that Baichuan-7B did not perform better than GPT-2.
This also indicates that the selective context method is not scalable.
This therefore indicates that the baseline we used is sufficiently robust.

\textbf{Retention Ratios}: In our experiments, we explore various content retention ratios: 0.2, 0.35, 0.5, 0.65, and 0.8. These ratios determine the proportion of content to be retained. This exploration allows us to examine the trade-off between efficiency and performance as the amount of retained information varies.

\textbf{Setting for Inference Measure:} To measure the inference acceleration and memory savings brought about by context compression, we conducted practical measurements using an NVIDIA 4090 GPU with 24GB of VRAM. The LLM model used was Baichuan-7B~\citep{Yang2023Baichuan2O}, and the data format employed was bfloat16. 

\section{Results and Discussions}
\subsection{Comparison to Original Context}

We initially compare the performance of Instruction-Aware Contextual Compressor with varying context retention ratios to the reranked original context, which utilizes the original context after reranking but no compression at all. All results are shown in table \ref{tab:perf_diff}, and the "diff" column represents the difference in performance compared to uncompressed text.

As shown in the table \ref{tab:perf_diff}, at retention rate of 0.8, the performance loss is minimal, with Rouge-1 showing only a marginal decrease in the range of 0.003 to 0.008. This demonstrates a high level of consistency between answers provided in compressed contexts and those in original contexts. Surprisingly, the Rouge-2 and Rouge-L score with the generation method is even higher than the original text, which was unexpected. This indicates that our method successfully filtered unrelated or even noisy content, improving the LLM's performance.

As the retention ratio decreases, the effectiveness of all methods declines, which is expected since there is less valuable information provided to the LLM. Overall, the generation method outperforms the ranking method, with a slower rate of performance decline from 0.8 to 0.35 compared to the ranking method. However, it was unexpected that at a retention ratio of 0.2, there was a sudden significant drop in performance, indicating a rapid loss of effectiveness for the generation method at low retention ratios.

In traditional machine learning, ensemble learning is widely regarded as a robust and effective method for improving model performance. Therefore, we propose using the "average rank" to combine the generation method and the ranking method. This involves taking the average of the ranks assigned by the generation method and the ranking method, resulting in a new ranking score for a given text. Overall, the "average rank" method outperforms both the generation and ranking methods. It shows significant improvement from 0.2 to 0.65 retention ratios, with fewer losses compared to the original context. 
At a retention ratio of 0.8, while it may not outperform the generation method, it still surpasses the ranking method.

\begin{table*}[ht]
\centering
\caption{Comparing Instruction-Aware Contextual Compressor with different context retention ratio to the original context}
\resizebox{\linewidth}{!}{
\begin{tabular}{lcccccc}
\hline
Method & Rouge-1 & Rouge-2 & Rouge-L & Rouge-1 Diff & Rouge-2 Diff & Rouge-L Diff \\
\hline
origin & 0.675 & 0.529 & 0.623 & 0 & 0 & 0 \\
ranking-0.8 & 0.667 & 0.522 & 0.617 & -0.008 & -0.007 & -0.006 \\
ranking-0.65 & 0.643 & 0.493 & 0.594 & -0.032 & -0.036 & -0.029 \\
ranking-0.5 & 0.633 & 0.481 & 0.584 & -0.042 & -0.048 & -0.039 \\
ranking-0.35 & 0.608 & 0.454 & 0.56 & -0.067 & -0.075 & -0.063 \\
ranking-0.2 & 0.587 & 0.429 & 0.539 & -0.088 & -0.1 & -0.084 \\
\hline
generation-0.8 & 0.672 & 0.53 & 0.624 & -0.003 & 0.001 & 0.001 \\
generation-0.65 & 0.66 & 0.515 & 0.611 & -0.015 & -0.014 & -0.012 \\
generation-0.5 & 0.642 & 0.495 & 0.596 & -0.033 & -0.034 & -0.027 \\
generation-0.35 & 0.62 & 0.471 & 0.574 & -0.055 & -0.058 & -0.049 \\
generation-0.2 & 0.559 & 0.4 & 0.509 & -0.116 & -0.129 & -0.114 \\
\hline
ensembled-0.8& 0.669 & 0.523 & 0.619 & -0.006 & -0.006 & -0.004 \\
ensembled-0.65& 0.66 & 0.515 & 0.611 & -0.015 & -0.014 & -0.012 \\
ensembled-0.5& 0.649 & 0.504 & 0.601 & -0.026 & -0.025 & -0.022 \\
ensembled-0.35& 0.628 & 0.479 & 0.582 & -0.047 & -0.05 & -0.041 \\
ensembled-0.2& 0.599 & 0.442 & 0.553 & -0.076 & -0.087 & -0.07 \\
\hline
\end{tabular}
}
\label{tab:perf_diff}
\end{table*}

\subsection{Comparison to Baseline}

\begin{figure}[h]
  \centering
  \includegraphics[width=0.6\linewidth]{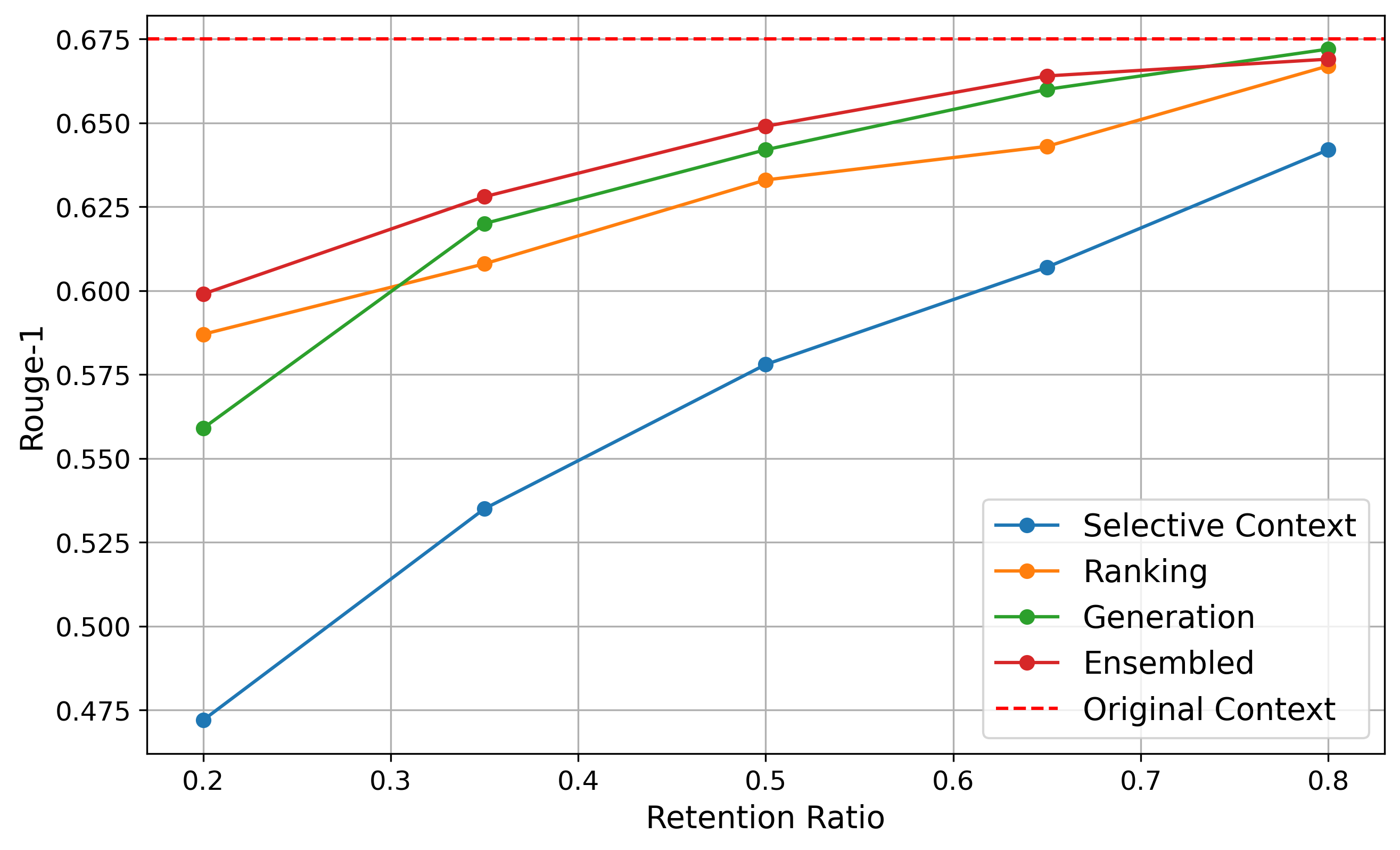}
  \includegraphics[width=0.6\linewidth]{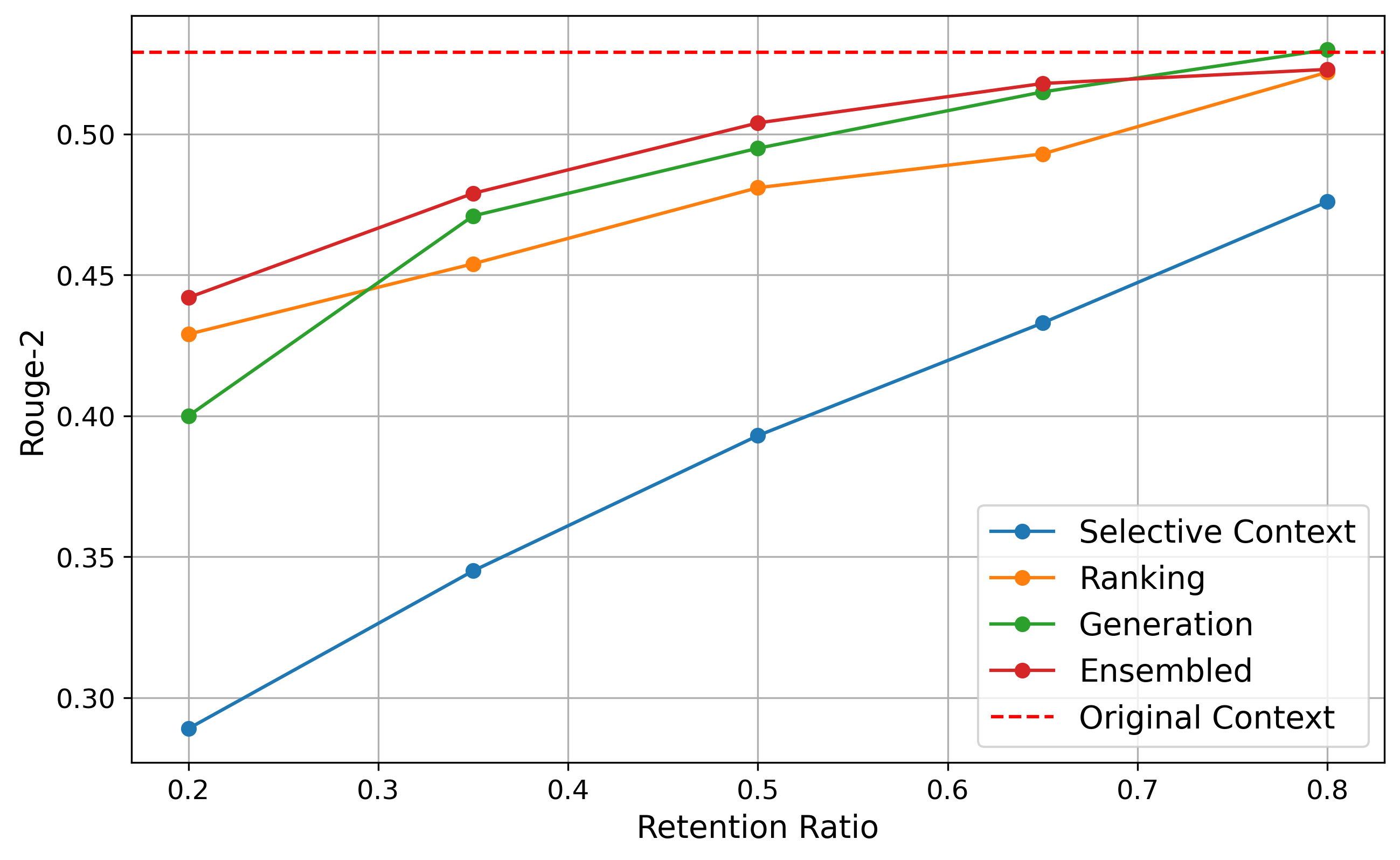}
  \includegraphics[width=0.6\linewidth]{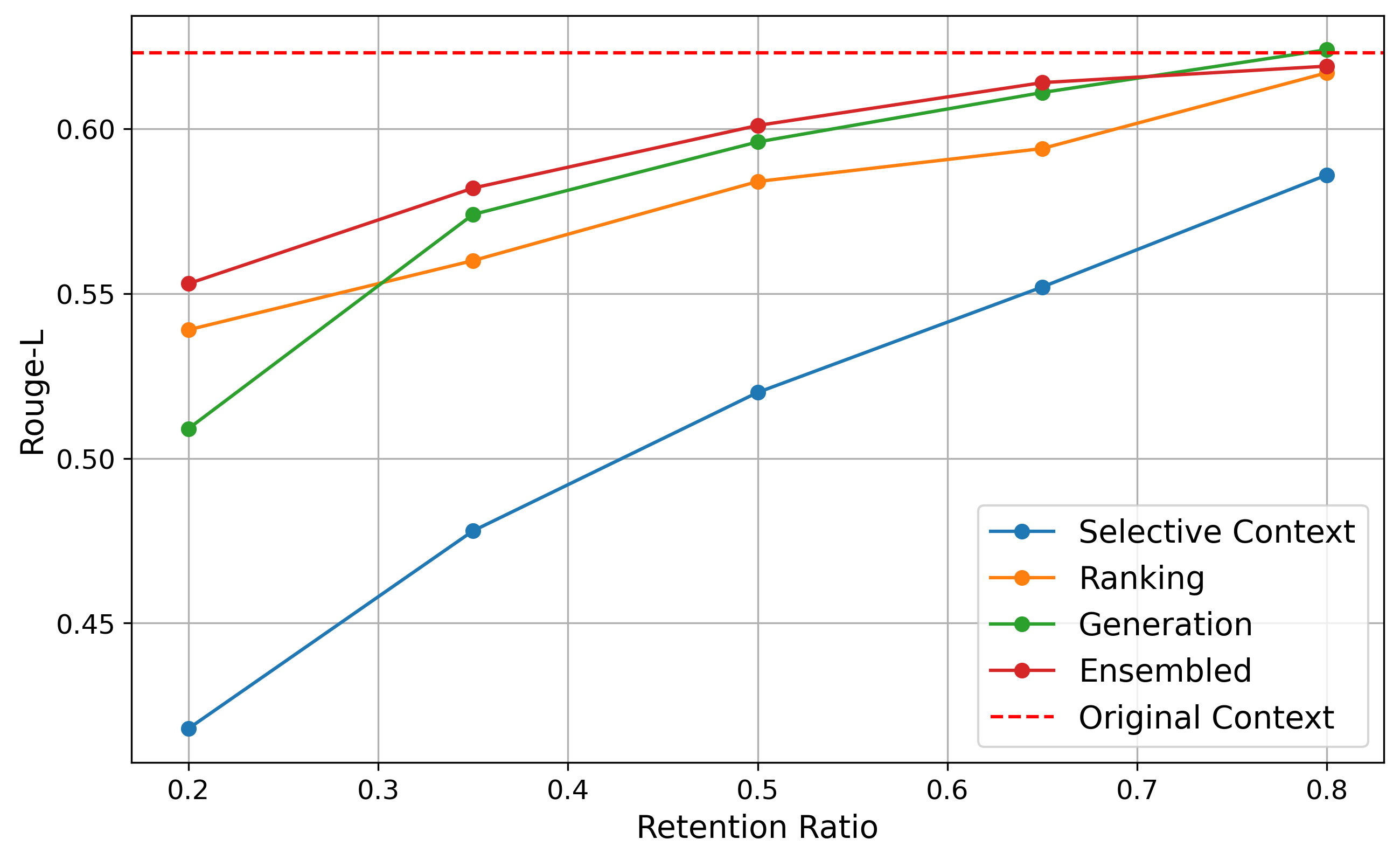}
  \caption{Performance of Instruction-Aware Contextual Compression compared to the Selective Context baseline}
  \label{fig:perf_vs_baseline}
\end{figure}

In this section, we compare our method to Selective Context baseline, and the results are presented in Figure \ref{fig:perf_vs_baseline}. 
Selective Context is a context compression method based on text self-information and represents the state-of-the-art as of the time of writing this paper. Comparing our method to Selective Context, which serves as a baseline, can effectively demonstrate the validity of our approach.
As shown in Figure \ref{fig:perf_vs_baseline}, our proposed method, Instruction-Aware Contextual Compressor, is even more effective compared to Selective Context. Both methods, whether purely ranking-based or generation-based, outperform Selective Context, and the lead becomes more significant as the retention ratio decreases. This indicates that our proposed method excels in selecting more informative content even when only a limited amount of information can be retained.

\subsection{The Impact of Generation Steps}

\begin{figure}[h]
  \centering
  \includegraphics[width=0.8\linewidth]{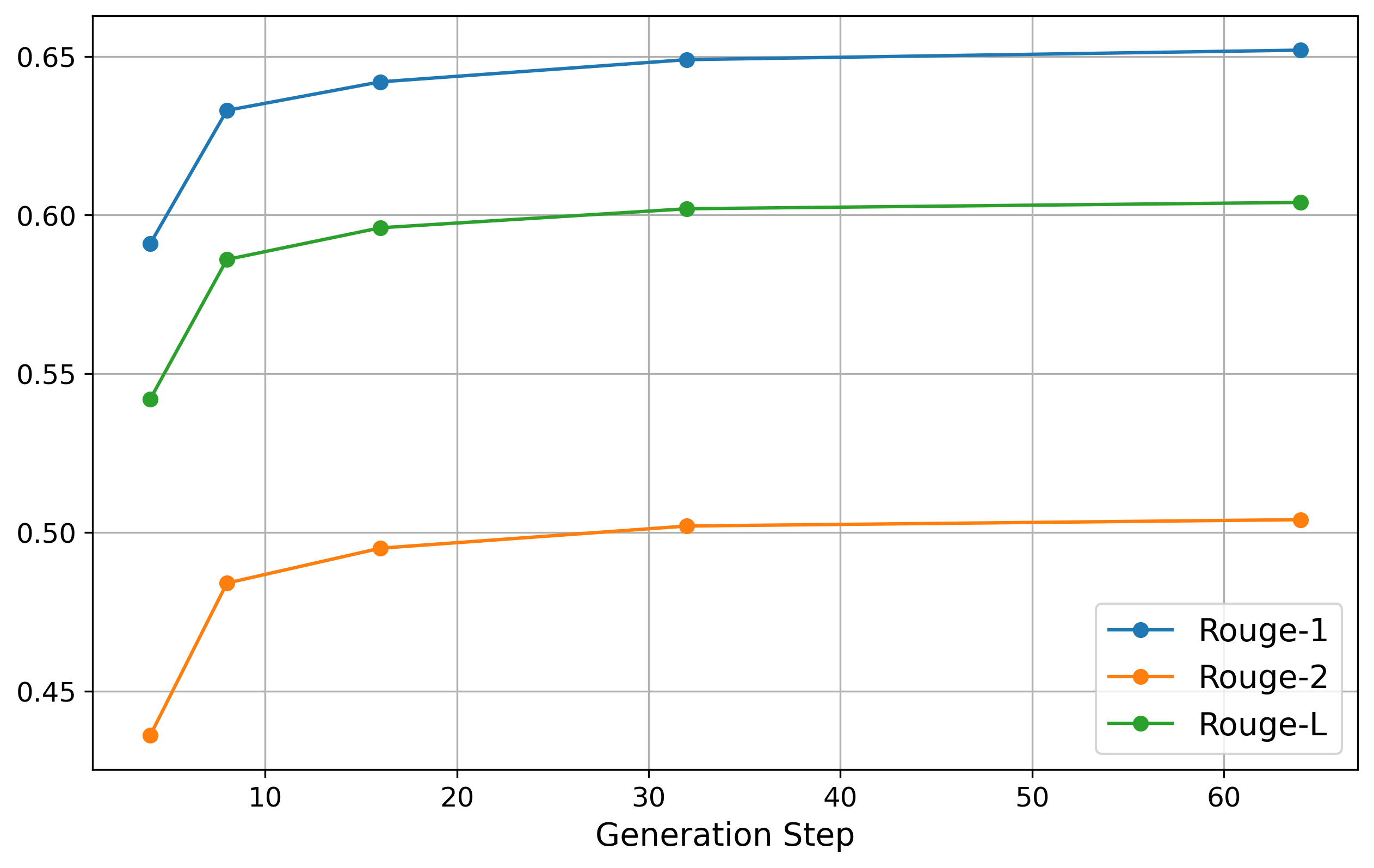}
  \caption{The Impact of Generation Steps on Context Compression Effectiveness}
  \label{fig:gen_step}
\end{figure}

For context compression using generative information, intuitively, if the number of generation steps is too few, it might not have generated a complete response. Consequently, the effectiveness at this stage may be suboptimal. As the number of generation steps increases, the effectiveness of compression is expected to improve. To explore this, we conducted experiments with a fixed retention ratio of 0.5, testing a series of generation step values ranging from 4 to 64. The results, as shown in Figure \ref{fig:gen_step}, indeed demonstrate that for generation-based context compression, the performance gradually improves with an increase in the number of steps. However, after reaching 32 steps, it reaches a plateau, indicating a diminishing marginal return with further increases in the number of generation steps.

\subsection{Speed Up and Memory Saving}
We also measured the impact of context compression on the Large Language Model (LLM). As demonstrated in the table \ref{tab:speedup}, when the retention ratio is set to 0.5, the inference speed per token increases by a factor of 2.2, while memory usage decreases by 5.05\%.

\begin{table}[htbp]
    \centering
    \caption{Speed up after context compression}
    \label{tab:speedup}
    \begin{tabular}{cccc}
        \toprule
        Retention Ratio & Ranking& Generation &Ensemble\\
        \midrule
        0.8 & 1.38& 1.00&0.94\\
        0.65 & 1.73& 1.17&1.08\\
        0.5 & 2.05& 1.31&1.20\\
        0.35 & 2.41& 1.45&1.32\\
        0.2 & 2.81& 1.58&1.43\\
        \bottomrule
    \end{tabular}
\end{table}

\subsection{Re-ranking Performance}

The ability to filter out irrelevant documents through re-ranking is also a crucial capability of Instruction-Aware Contextual Compression. Following a hierarchical design approach, a large number of initially retrieved documents are first re-ordered using Instruction-Aware Contextual Compressor to select the top-k documents, which are then further compressed. Therefore, we present the retrieval performance of Instruction-Aware Contextual Compressor on extensive datasets, measured by Recall@1, 5, and 10, as shown in \ref{tab:recall}.

\begin{table}[htb]
    \caption{Re-ranking Performance of Instruction-Aware Contextual Compressor}
    \label{tab:recall}
    \begin{tabular}{@{}clccc@{}}
        \toprule
        &Dataset     &Recall@1   &Recall@5  &Recall@10  \\ 
        \midrule
        & wikipedia-cn-20230720-dataset   &0.975   &0.998  &0.999  \\
        & wiki\_atomic\_edits & 0.968 & 0.997  & 0.999  \\
        & alpaca\_gpt4   & 0.789 & 0.933  & 0.971 \\ 
        & bq         &0.718  & 0.833 & 0.903 \\
        & firefly   &0.632 &0.849 &0.947 \\
        & webqa   &0.667 &0.913 &0.974 \\
        & dureader\_dataset   &0.913 &0.988 &0.996 \\
        & cmrc2018   &0.972 &0.986 & 0.993\\
        & csl  &0.846 &0.961 &0.986 \\
        & pawsx  &0.606 &0.996 &1.000 \\
        & dureader\_robust  & 0.627 &0.859 &0.921 \\
        & $\rm T^2$Ranking\_train\_dataset  &0.405 &0.756 &0.901 \\
        &tiracl   &0.727 &0.891 &0.956 \\
        & belle\_2m  &0.749 &0.921 &0.972 \\
        & mlqa  &0.718 &0.908 &0.960 \\
        & lcqmc  &0.435 &0.876 &0.960 \\
        & hc3\_chinese  &0.270 &0.751 & 0.897\\
        & zhihu\_kol  &0.238 &0.557 &0.787 \\
        & xlsum  &0.395 &0.833 &  0.933\\
        & ocnli  &0.371 &0.876 &0.961 \\
        & chatmed\_consult  &0.594 &0.803 &0.899 \\
        \bottomrule
    \end{tabular}
\end{table}

\section{Conclusion}

In this paper, we introduced Instruction-Aware Contextual Compression to filter out less relevant content, providing a more concise and efficient context representation for LLMs, all without compromising their performance. 
An important discovery we made is that generation-based Instruction-Aware Contextual Compression is more effective than ranking-based Instruction-Aware Contextual Compression methods. With generation-based Instruction-Aware Contextual Compression, using only 1/10 of the data, the results can surpass those of the Instruction-Aware Contextual Compression method. 
With only 50\% of the context retained, we achieved a 2.2x inference speedup for the LLM and saved 5\% of GPU VRAM, while the Rouge-1 metric only dropped by 0.047.
According to our evaluations, the results show that Instruction-Aware Contextual Compressor significantly improves the efficiency of LLMs and serves as a valuable component in the Retrieval-augmented Generation pipeline.

\section*{Data availability and access}
The code and data used in this project are open-sourced and available at \href{https://github.com/howard-hou/instruction-aware-contextual-compressor}{https://github.com/howard-hou/instruction-aware-contextual-compressor}. The provided information is sufficient to reproduce the results of this paper. Additional data supporting the findings of this study can be obtained from the corresponding author upon reasonable request.

\newpage

\bibliography{sn-bibliography}% common bib file

\section*{Acknowledgements}
This project was supported by the Industrial Metaverse project of Guangdong Laboratory of Artificial Intelligence and Digital Economy (SZ), with the project number 000015; it was also supported by the Guangming District Government GPT Service project, with the project number 23210016.

\section*{Author information}
Haowen Hou and Fei Ma are contributed equally to this work.

\subsection*{Authors and Affiliations}
Guangdong Laboratory of Artificial Intelligence and Digital Economy(SZ), Yutang, Shenzhen, 518000, Guangdong, China.

Haowen Hou, Fei Ma, Binwen Bai, Xinxin Zhu \& Fei Yu

\subsection*{Contributions}
All authors have contributed to the ideas and design of this research.
Haowen Hou was in charge of developing and training the model, and also wrote this manuscript.
Fei Ma carried out the experiments and the analysis of the outcomes.
Binwen Bai and Xinxin Zhu managed the data collection, arrangement, cleansing, and its ultimate transformation into the training format.
Fei Yu offered feedback on the preliminary drafts of the manuscript.
All authors have reviewed and approved the final manuscript.

\subsection*{Corresponding author}
Correspondence to Haowen Hou.

\section*{Ethics declarations}
\subsection*{Competing interests}
The authors confirm that there are no financial conflicts or personal connections that might be perceived as affecting the findings presented in this study.

\subsection*{Ethical and informed consent for data used}
There are no concerns regarding the ethical use of the data or the informed consent process. All necessary ethical guidelines and protocols have been followed, and informed consent was obtained from all participants involved in the study.

\end{document}